\documentclass[a4paper]{article}

\usepackage{INTERSPEECH2021}
\usepackage{comment}
\usepackage{placeins}
\usepackage{xcolor}
\usepackage{float}
\usepackage{makecell}
 \usepackage{url}

\makeatletter
\renewcommand{\section}{\@startsection
  {section}%
  {1}%
  {}%
  {0.6\baselineskip}%
  {0.6\baselineskip}%
  {}}%

\renewcommand{\subsection}{\@startsection
  {subsection}%
  {2}%
  {}%
  {0.1\baselineskip}%
  {0.1\baselineskip}%
  {}}%

\renewcommand{\subsubsection}{\@startsection
  {subsubsection}%
  {3}%
  {}%
  {0.1\baselineskip}%
  {0.1\baselineskip}%
  {}}%

\g@addto@macro\normalsize{%
  \setlength\abovedisplayskip{5pt plus 2pt minus 2pt}
  \setlength\belowdisplayskip{5pt plus 2pt minus 2pt}
  \setlength\abovedisplayshortskip{4pt plus 2pt minus 2pt}
  \setlength\belowdisplayshortskip{4pt plus 2pt minus 2pt}
}

\makeatother

\title{Alzheimer’s Disease Detection from Spontaneous Speech through Combining Linguistic Complexity and (Dis)Fluency Features with Pretrained Language Models}
\name{Yu Qiao$^1$, Xuefeng Yin$^1$, Daniel Wiechmann$^2$, Elma Kerz$^1$}
\address{
  $^1$RWTH Aachen University\\
  $^2$University of Amsterdam}
\email{yu.qiao@rwth-aachen.de, xuefeng.yin@rwth-aachen.de, d.wiechmann@uva.nl, elma.kerz@ifaar.rwth-aachen.de}

\begin{document}

\maketitle
\begin{abstract}
\bigskip
In this paper, we combined linguistic complexity and (dis)fluency features with pretrained language models for the task of Alzheimer's disease detection of the 2021 ADReSSo (Alzheimer’s Dementia Recognition through Spontaneous Speech) challenge. An accuracy of 83.1\% was achieved on the test set, which amounts to an improvement of 4.23\% over the baseline model. Our best-performing model that integrated component models using a stacking ensemble technique performed equally well on cross-validation and test data, indicating that it is robust against overfitting.

\end{abstract}

\noindent\textbf{Index Terms}: Alzheimer’s disease, disfluency, pretrained language models, automated Alzheimer’s disease detection, linguistic complexity

\section{Introduction}

Alzheimer's disease (AD) is a gradual and progressive neurodegenerative disease caused by neuronal cell death \cite{mattson2004pathways}. The number of people diagnosed with AD is rapidly increasing\footnote{https://www.alz.org/alzheimers-dementia/facts-figures}. The high prevalence of the disease and the high costs associated with traditional approaches to detection make research on automatic detection of AD critical \cite{zeisel2020world}. A growing body of research has demonstrated that quantifiable indicators of cognitive decline associated with AD are detectable in spontaneous speech (see \cite{de2020artificial} for a recent review). These indicators encompass acoustic features, such as vocalisation features (i.e. speech-silence patterns) \cite{luz2017longitudinal}, paralinguistic features, such as fluency features \cite{campbell2021paralinguistic} and speech pause distributions \cite{pastoriza2021speech}, as well as syntactic and lexical features extracted from speech transcripts \cite{bucks2000analysis}.

This area of research has benefited from recent advances in natural language processing (NLP) and machine learning, as well as an increasing number of interdisciplinary research collaborations. A prime example of this is the ADReSS(o) (Alzheimer's Dementia Recognition through Spontaneous Speech) Challenge, aimed at generating systematic evidence for the use of such indicators in automated AD detection systems and towards their clinical implementation. This challenge has made significant contributions to research on AD detection by enabling the research community to test their existing methods, develop novel approaches and to benchmark their AD detection systems on a shared dataset. The ADReSSo Challenge at INTERSPEECH 2021 \cite{luz2021detecting} is geared towards automatic recognition of AD from spontaneous speech and involved three subtasks. Here in this paper, we focus on the AD classification subtask, for which research teams were asked to build a model to predict the label (AD or non-AD) for a short speech session. Participating teams could use the speech signal directly and extract acoustic features or automatically convert the speech to text (ASR) and extract linguistic features from this ASR-generated transcript.

\subsection{Related work}
 
In this section, we provide a concise review of research on automatic AD detection through speech, with particular attention to previous studies conducted as part of the 2020 ADReSS Challenge. The AD classification approaches in this challenge relied on a wide range of acoustic, paralinguistic, and linguistic features or their combination.  Classification accuracy scores of the proposed models ranged between 68\% and 89.6\%. While some approaches either focused on acoustic or linguistic features, the best performing contributions in the 2020 challenge embraced a multi-modal approach combining several types of features (e.g. \cite{yuan2020disfluencies}\cite{syed2020automated}\cite{balagopalan2020bert}). Furthermore, building on earlier work reporting on the effectiveness of the use of word embeddings in AD detection (\cite{guerrero2020word}\cite{mirheidari2018detecting}), several approaches successfully employed pretrained language models (e.g.  \cite{yuan2020disfluencies}\cite{syed2020automated}\cite{balagopalan2020bert}). Another important issue addressed in several studies concerned how to deal with variance in the predictive performance of pretrained models resulting from fine-tuning for downstream tasks with a small data set. In response to this issue, the authors of the best performing model \cite{yuan2020disfluencies} introduced an ensemble method to increase the robustness of their approach.
In response to this issue, the best performing paper of the 2020 challenge \cite{yuan2020disfluencies} introduced an ensemble approach to increase the robustness of their models. Finally, it is important to note that some of the high-performing models in last year's challenge -- including the best model described in \cite{yuan2020disfluencies} -- used rich manual transcription that included pause and disfluency annotation. Such transcripts were not provided in the 2021 challenge, making it more demanding compared to last year's challenge. 

\subsection{Modeling approach}

The modeling approach presented in this paper builds on key insights reported in the studies reviewed above and extends on these (1) by integrating linguistic indicators of linguistic complexity and sophistication, features of (dis)fluency and transformer-based pretrained language models and (2) by utilizing ensembling methods to combine the information from these feature groups and to reduce the variance in model predictions. Specifically, we perform experiments with classification based on three ensembling techniques: Ensembling by bagging via majority vote, ensembling by bagging using feature fusion, and ensembling by stacking.


\section{Data and analysis}

\subsection{Data}

The Alzheimer's Disease Detection dataset provided by the organizers of the ADReSSo Challenge 2021 consists of speech recordings of picture descriptions from the Boston Diagnostic Aphasia Exam produced by 87 individuals with an AD diagnosis and 79 cognitively normal subjects (control group).  The recordings were acoustically enhanced (noise reduction through spectral subtraction) and normalised. The data were also balanced with respect to age and gender. Besides the audio files, the organizers provided segmentations of the recordings into vocalisation sequences with speaker identifiers. No transcripts were provided.

\subsection{Speech Recognition}

We used AppTek's Automatic Speech Recognition technology via a cloud API service\footnote{https://www.apptek.com/} for automatically transcribing the audio files. 
The transcripts were converted from XML into raw text formats with full stops being added at the end of each utterance based on the segmentations provided by the organizers. These files served as the input for the automated text analysis (see Section 2.4).

\subsection{(Dis)fluency}

To model the speakers' articulatory  (in particular (dis)fluency-related) characteristics, we derived several features from the ASR system that fall into four classes. (1) \textit{Silent pauses} - The ASR output contained the start- and end-times as well as confidence scored for each recognized word. Durations of pauses were calculated from forced alignment and binned by duration into short pauses ($<2 sec$) and long pauses ($>2 sec$). In addition, we calculated the total pause duration per sentence (in seconds). (2) \textit{Speed of articulation} - We enriched the output of the ASR with syllable counts from the Carnegie Mellon University Pronouncing Dictionary\footnote{http://www.speech.cs.cmu.edu/cgi-bin/cmudict}. Based on this information we assessed the mean syllable duration as well as syllables per minute for each utterance in the speech data. (3) \textit{Filled pauses} - Next to the number and total duration of silent pauses, we derived frequency counts per sentence for two filled pause type, \textit{uh} and \textit{um}, that had been shown to discriminate between AD patients and controls in previous studies \cite{yuan2020disfluencies}. (4) \textit{Pronunciation} - As the known symptoms of AD patients include mispronunciation \cite{orange1996conversational}, we calculated average word level confidence scores as a proxy of pronunciation quality, which have been employed for the speech pattern detection in the context of detection of Alzheimer’s Disease \cite{pan2020improving}. All measures were calculated at utterance level. An overview of these measures with descriptive statistics for AD patients and control subjects is presented in Table 1.

 \begin{table}[t]
     \centering
       \setlength{\tabcolsep}{2pt}
     \caption{Descriptive statistics of (dis)fluency measures}
     \begin{tabular}{l|cc|cc}
     \hline
     &\multicolumn{2}{c|}{AD patients}&\multicolumn{2}{c}{Control}\\
          (Dis)Fluency measure &M &SD &M &SD\\
\hline
\textit{Speed of articulation}&&&&\\
Mean syllable duration&0.28&0.05&0.26&0.03\\
Syllables per minute&205&45.7&224&35.7\\
\hline
\textit{Silent pauses}&&&&\\
Pause time per sentence (in sec)&0.92&0.89&0.63&0.49\\
N long pauses ($>2 sec$)&1.28&2.22&0.473&0.71\\
N short pauses ($<2 sec$)&13.2&9.28&15.4&11.7\\
\hline
\textit{Filled pauses}&&&&\\
N \textit{uh}&0.29&0.88&0.24&0.54\\
N \textit{um}&0.07&0.37&0.31&0.74\\
\hline
\textit{Pronunciation}&&&&\\
Mean ASR confidence &0.83&0.09&0.86&0.08\\
\hline
     \end{tabular}
     
     \label{tab:my_label}
     \vspace{-5mm}
 \end{table}


\subsection{Automated Text Analysis (ATA)}

\begin{figure}[b]
    \centering
    \includegraphics[width= 0.5\textwidth]{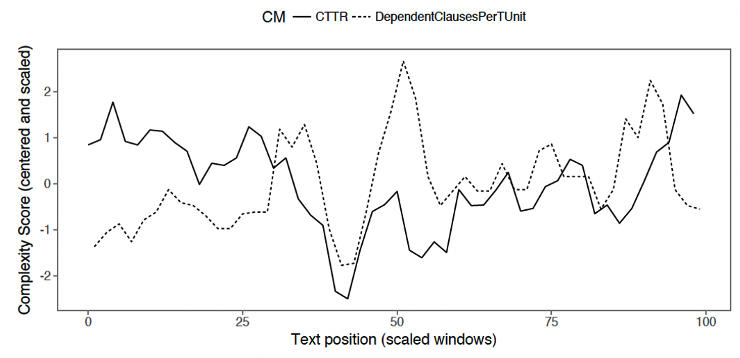}
    \caption{Schematic representation of `complexity contours' for two out of 293 complexity measures (CM) investigated: CTTR (Corrected Type Token Ratio) and Dependent Clauses per TUnit). Centering/scaling was applied here only for purposes of illustration.}
    \label{fig:coco}
    \vspace{-5mm}
\end{figure}

The speech transcripts were automatically analyzed using CoCoGen (short for: Complexity Contour Generator), a computational tool that implements a sliding window technique to calculate within-text distributions of scores for a given language feature (for current applications of the tool in the context of text classification, see \cite{kerz2020becoming,qiao2020language,strobel2020relationship}). In this paper, we employed a total of $293$ features derived from interdisciplinary, integrated approaches to language \cite{Christiansen2017} that fall into four categories:  (1) measures of syntactic complexity, (2) measures of lexical richness, (3) register-based n-gram frequency measures, and (4) information-theoretic measures. In contrast to the standard approach implemented in other software for automated text analysis that relies on aggregate scores representing the average value of a feature in a text, the sliding-window approach employed in CoCoGen tracks the distribution of the feature scores within a text.  A sliding window can be conceived of as a window of size $ws$, which is defined by the number of sentences it contains. The window is moved across a text sentence-by-sentence, computing one value per window for a given indicator. In the present study, the $ws$ was set to 1. The series of measurements generated by CoCoGen captures the progression of language performance within a text for a given indicator and is referred here to as a `complexity contour' (see Figure  \ref{fig:coco} for illustration). CoCoGen uses the Stanford CoreNLP suite \cite{manning2014stanford} for performing tokenization, sentence splitting, part-of-speech tagging, lemmatization and syntactic parsing (Probabilistic Context Free Grammar Parser \cite{klein2003accurate}).

\subsection{Pretrained Language Models}

Since their inception, transformer-based pretrained language models such as BERT \cite{devlin2018bert} and ERNIE \cite{sun2019ernie20} have achieved state- of-the-art performance
in various classification tasks.
The results of previous research demonstrate that the language characteristics of AD too can be captured by pretrained language models fine-tuned to the task of AD classification (see above). In this paper, pretrained BERT and ERNIE models were fine-tuned for the AD classification task and combined with classifiers trained on complexity and (dis)fluency features (see Section 3). Each of the 161 speakers in the training data is considered as a data point. The input of the model consists of all the text sequences of each speaker obtained by the ASR system, and the output is the class of the corresponding speaker, 0 for Control and 1 for AD.

\section{Experimental Setup}

In this section we describe the component models used in our approach and how they were combined.
To assess the performance of each model, 5-fold cross validation was used.

\subsection{CNN Complexity + (Dis)Fluency Models}
In order to make optimal use of the complexity and (dis)fluency features, which are sequential in nature, we built convolutional neural network (CNN) models. Originally proposed in computer vision, CNNs have been successfully adapted to various NLP tasks \cite{collobert2011natural} and sentence classification tasks \cite{kalchbrenner2014convolutional}\cite{kim2014convolutional}\cite{ma2015dependency}. The CNN model has the advantage over models that rely on aggregated features, e.g. mean feature values, in that it is capable of capturing patterns in a feature sequence. We followed the approach proposed by \cite{kim2014convolutional}, but replaced the word embedding with the concatenation of complexity and (dis)fluency features. Due to the small size of the dataset, we set the size of filters to be $2\times d$, $3\times d$, $4\times d$  where $d$ is the input feature dimension. Eight filters were used for each of the three filter types.

\subsection{Fine-tuned BERT and ERNIE Models}
The Huggingface Transformers library \cite{wolf-etal-2020-transformers} was adopted for fine tuning pretrained language models. Bert-for-Sequence-Classification was used and initialized with `bert-base-uncased' and `nghuyong/ernie-2.0-en' as our pretrained BERT and ERNIE model, respectively. In both cases, the base model was used rather than the large one, as preliminary experiments revealed no reliable differences in terms of classification accuracy between the two models on our dataset. Both models consists of 12 Transformer layers with hidden size 768 and 12 attention heads. The following hyperparameters were used for fine-tuning: the learning rate was set to $2\times10^{-5}$ with 50 warmup steps and $l_2$ regularization set to 0.1. The maximum sequence length for both models was set to 256. For both models, default tokenizers were used.

\subsection{Use of Ensembling Methods}

Previous research on predicting AD using pretrained language models has demonstrated that their predictions based on fine-tuning for downstream tasks with a small dataset tend to be brittle and subject to high variance. To reduce this variance, we used an adapted version of the ensembling approach proposed in \cite{yuan2020disfluencies}: Each of the models described above was trained $50$ times ($N=50$). During the prediction phase, each model instance independently generated a prediction. The final classification decision was then determined by hard-voting, i.e.  each model contributed its class prediction as a vote and the class that receives the majority of the votes was returned by the ensemble model. Besides using ensemble methods so as to reduce the variance in the prediction of a model, we also employed them to integrate information from different models. To this end, we performed experiments with two types of ensemble based methods, which are referred to here as \textit{ensembling by bagging} and \textit{ensembling by stacking}. Bagging involves fitting several independent models and pooling their predictions in order to obtain a model with a lower variance, while stacking involves combining the models by training a meta-model to output a prediction based on the different models predictions (see below). In each of the combined models, we used the same hyperparameter settings as stated above.
\begin{figure}[!htp]
    \centering
    \includegraphics[width = 0.33\textwidth]{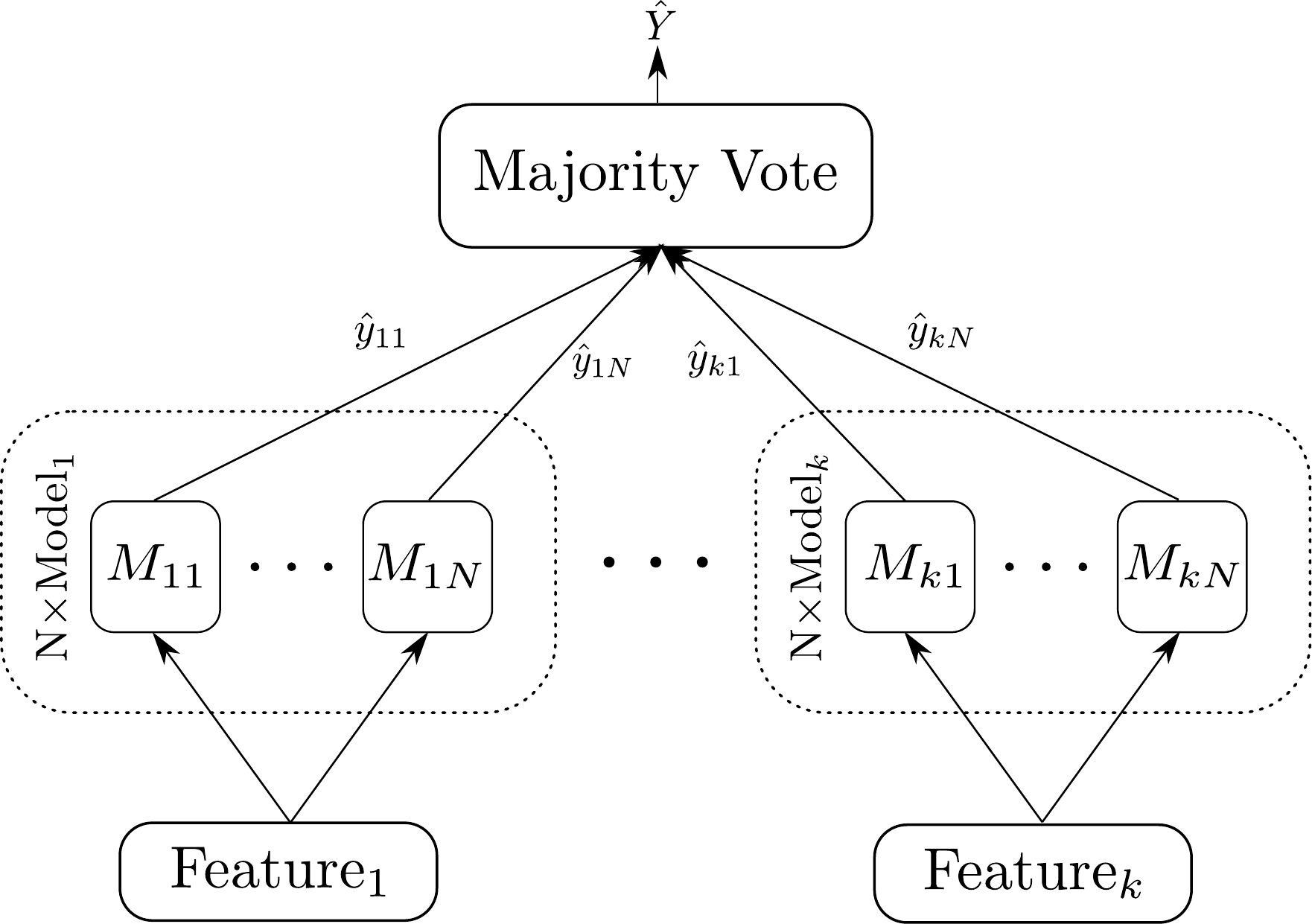}
    \caption{Structure diagram of Model A. During training, we train each of the $k$ models $N$ times. During inference, $j$th instance of model $i$ gives prediction $\hat{y}_{ij} $ independently. The final output of the ensembled model $\hat{Y}$ is the label, which the majority of the $k\times N$ model instances agree upon.}
    \label{fig:majority_vote}
    \vspace{-3mm}
\end{figure}

\subsubsection{Model A: Ensembling by bagging via majority vote}
    Ensembling by bagging via majority vote has been shown to be a simple yet effective method to increase the performance of classification models \cite{costello2018multilayer}\cite{oza2008classifier}. The first classification model (Model A) employed majority voting among 50 CNNs that used complexity and (dis)fluency features and 50 ERNIE models (see Figure \ref{fig:majority_vote}). That is, as specified above, in this approach, each model was first trained/fine tuned $50$ times, meaning that the final classification was based on 100 model instances. The classification in the Model A approach was then determined by counting the votes for each class (AD and controls (CN)) and choosing the more frequent class as the predicted one.

\begin{figure}[ht]
    \centering
    \includegraphics[width = 0.3\textwidth]{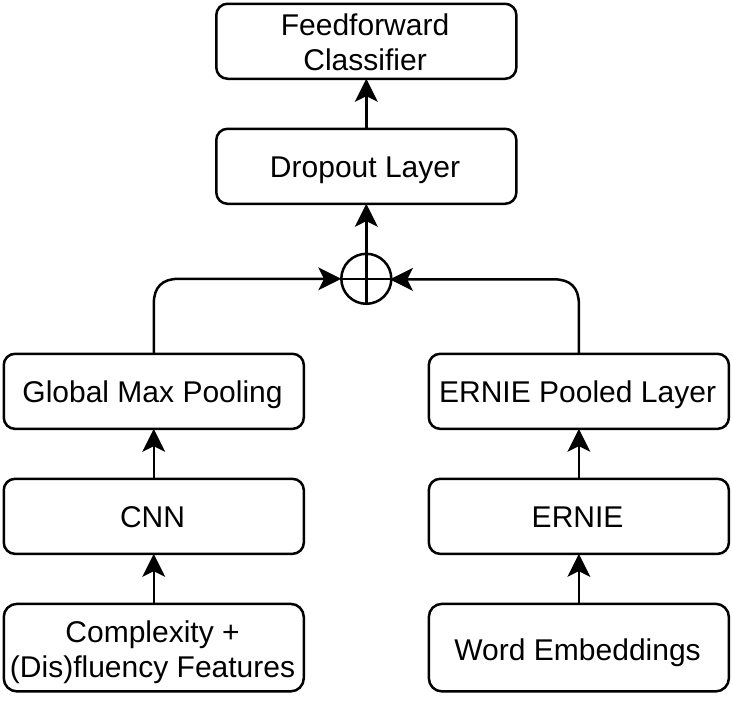}
    \caption{Structure diagram of Model B.}
    \label{fig:hybrid}
    \vspace{-4mm}
\end{figure}
\subsubsection{Model B: Ensembling by bagging using feature fusion}
The second model (Model B) combined a CNN and a BERT model, which has previously been shown to perform better than either model alone \cite{alghanmi2020combining}. Following the approach taken by \cite{alghanmi2020combining}, we built a model in which complexity and (dis)fluency information was first concatenated at the feature-level and subsequently fed into a CNN (see Figure \ref{fig:hybrid}. 
The hidden vector coming from this CNN is then concatenated with ERNIE. More specifically, the pooled output vector of the [CLS]\footnote{[CLS], stands for classification, is a special
token added in front of every input samples of BERT/ERNIE model to represent sample-level classification \cite{devlin2018bert}.} token for Ernie model. This concatenated hidden vector will serve as the input of a feed forward classifier on top of CNN and Ernie. To train this model, we first fine-tune ERNIE model. Then we freeze the parameters of the fine-tuned ERNIE model and jointly train the CNN model and combined feedforward classifier.

\begin{table*}[htb!]
\centering
\caption{Mean accuracy (with standard deviations), precision, recall and F1 scores over a 5 fold cross-validation}
\setlength{\tabcolsep}{3pt}
\begin{tabular}{|l|l|l|l|l|l|l|l|}
\cline{3-8}
                     \multicolumn{2}{c|}{}            & \multicolumn{2}{c|}{Precision} & \multicolumn{2}{c|}{Recall} & \multicolumn{2}{c|}{F1} \\ \hline
Model & Acc & CN             & AD            & CN           & AD           & CN         & AD         \\ \hline
CNN Comp & M (SD)    &       M (SD)         &    M (SD)           &   M (SD)           &  M (SD)            &         M (SD)   &     M (SD)       \\ \hline
CNN[Comp+DisFl] &  0.80(0.06)   &     0.79(0.06)           &          0.81(0.08)     &     0.78(0.07)         &      0.83(0.06)        &     0.78(0.05)       &     0.82(0.07)\\ 
Bert-Base     & 0.79(0.06)   & 0.77(0.09)              & 0.84(0.08)              &  0.81(0.11)            & 0.78(0.12)             & 0.78(0.06)           & 0.80(0.07)            \\ 
Ernie-Base     & 0.80(0.04)     & 0.80(0.08)                &0.81(0.04)               &0.77(0.07)              &0.83(0.09)              &0.78(0.04)            &0.82(0.05)            \\ \hline\hline
\makecell[l]{\textbf{Model A:} CNN[Comp+DisFl]+[Ernie] \\(\textit{sep mod, bagging})}     & 0.76(0.07)    &        0.61(0.13)        &      0.88(0.05)         &    0.79(0.08)          &    0.74(0.08)          &     0.68(0.10)       &     0.80(0.06)       \\ 
\makecell[l]{\textbf{Model B:} CNN[Comp+DisFl]+[Ernie] \\(\textit{fusion, bagging})}     &   0.83(0.06)  & 0.75(0.11)               &      0.89(0.04)         &       0.83(0.09)       &     0.82(0.06)         &       0.78(0.09)     &     0.85(0.04)       \\ 
\makecell[l]{\textbf{Model C:} LR[Comp]+LR[DisFl]+[Ernie]+[Bert] \\(\textit{stacking})}     &\textbf{0.83}(0.07)    &0.82(0.10)                &0.85(0.09)               &0.83(0.10)              &0.84(0.09)              &0.82(0.08)            &0.84(0.07)            \\ \hline
\end{tabular}
\label{tab:training}
\vspace{-3mm}
\end{table*}







\begin{figure}
    \centering
    \includegraphics[width = 0.478\textwidth]{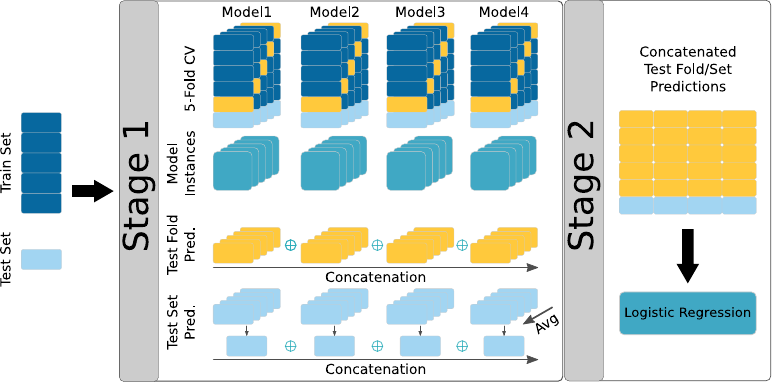}
    \caption{Schematic representation of ensembling by stacking.}
    \label{fig:modelC}
    \vspace{-3mm}
\end{figure}
\vspace{-0pt}

\begin{table*}[h]
\centering
\caption{Performance of the three ensemble models on test set}
\begin{tabular}{ll|l|l|l|l|l|l|}
\cline{3-8}
                            &     & \multicolumn{2}{c|}{Precision} & \multicolumn{2}{c|}{Recall} & \multicolumn{2}{c|}{F1} \\ \hline
\multicolumn{1}{|l|}{Model} & Acc & CN             & AD            & CN           & AD           & CN         & AD         \\ \hline
\multicolumn{1}{|l|}{\textbf{Model A: }CNN[Comp+DisFl]+Ernie(\textit{sep mod, bagging})}     &  0.79  &        0.77  &    0.81      &    0.83      &     0.74     &  0.80&    0.78       \\ \hline
\multicolumn{1}{|l|}{\textbf{Model B:} CNN[Comp+DisFl]+Ernie (\textit{fusion, bagging})}     & 0.75    &     0.73           &     0.77          &   0.81           &   0.69           &    0.76        &     0.72       \\ \hline
\multicolumn{1}{|l|}{\textbf{Model C:} LR[Comp]+LR[DisFl]+Ernie+Bert (\textit{stacking})}     & \textbf{0.83 }   & 0.82                &  0.85           &    0.86         &     0.80         & 0.84         &      0.82    \\ \hline
\end{tabular}
  \label{tab:testing}
  \vspace{-4mm}
\end{table*}
\subsubsection{Model C: Ensembling by stacking}
The final model, Model C, used in our experiments employed a stacking approach to ensemble all models \cite{wolpert1992stacked}, which has been shown to effectively increase the accuracy of the ensembled individual models. Specifically, we employed model stacking to combine two logistic regression models (LR) using complexity and (dis)fluency features respectively, and the two pretrained language models, i.e. BERT and ERNIE. 
The training procedure consists of two stages (see Figure \ref{fig:modelC}). First, in stage one, each of the four models is trained/fine-tuned independently using 5-fold cross-validation (CV). For each sample in the test fold, we obtain one prediction vector from each of the four models (Models 1 to 4). These predictions vectors are then concatenated and constitute the input data in a subsequent stage (stage 2). The final predictions of Model C are derived from another logistic regression model trained on the concatenated prediction vectors from stage 1. To perform inference on the test set, we take the predictions from all model instances trained in stage 1 and average them by model, which will served as input of stage 2 after concatenation. All hyperparameters for the training/fine-tuning of each of the ensembled models were selected as specified above.

\vspace{-5pt}
\section{Evaluation}
\vspace{-5pt}
In this section, we present our results on the AD detection task. The evaluation metrics for detection (accuracy, precision, recall, and F1 score) on the cross-validation (CV) set are presented in Table \ref{tab:training}. The results on the evaluation set are shown in Table \ref{tab:testing}. As indicated by boldface numbers, the best performing model in both cross-validation (mean accuracy = 83.16\%) and testing (accuracy = 83.10\%) was Model C, i.e. the model that combined complexity and (dis)fluency features with both pretrained language models using stacking. Model B, which combined a CNN trained on utterance-level complexity and (dis)fluency features with the best performing fine-tuned pretrained language model (ERNIE) using late fusion and ensembling by bagging, fell close behind reaching 82.7\% accuracy in CV. Model A, which combined the same features using majority voting with separate classifiers, performed below the accuracy levels of its component models, reaching 75.69\% accuracy in CV. On the test set, the accuracy score of 83.1\% of the best performing model, Model C, constitutes an improvement by 4.23\% over the baseline model, which was based on fusion of linguistic and acoustic features \cite{luz2021detecting}. Surprisingly, the relative performances of Model A and Model B were reversed on the test set, with Model A matching the performance of the baseline exactly (accuracy = 78.87\%) and Model B falling just short of that (accuracy = 74.65\%). The considerable discrepancies between the CV and test set classification accuracy for these models suggest that they suffer from overfitting. In contrast, Model C, which employed the stacking technique, performed equally well on CV and test data, indicating that it is robust against overfitting.

\vspace{-5pt}
\section{Discussion and Conclusion}
\vspace{-5pt}
The work presented here combined linguistic complexity and (dis)fluency features with pretrained language models for the task of Alzheimer's disease detection.  An accuracy of 83.1\% was achieved on the test set, which amounts to an improvement of 4.23 \% over the baseline model, which was based on fusion of linguistic and acoustic features. Our best performing model combined component models using a stacking ensemble technique. A key finding of this study is that incorporating information on linguistic complexity and (dis)fluency improved the performance of fine-tuned pretrained language models in AD classification by 3\%, suggesting that different component models encode complementary information regarding the characteristic language patterns of AD. Another important aspect of our results is that the ensemble model trained on `complexity contours', i.e. utterance-level measurements of human-interpretable complexity and fluency features, was able to match the performance of both fine-tuned pretrained BERT-like language models: Using 5-fold cross-validation with ensembling of 50 models in each fold, we obtained robust performance scores ($\approx 80\%$) for both types of models. 
This finding has important implications in light of increasing calls for moving away from black-box models towards white-box (interpretable) models for critical industries such as healthcare, finances and news industry \cite{rudin2019stop,loyola2019black}.


\bibliographystyle{IEEEtran}

\bibliography{mybib}

\begin{thebibliography}{10}
\providecommand{\url}[1]{#1}
\csname url@samestyle\endcsname
\providecommand{\newblock}{\relax}
\providecommand{\bibinfo}[2]{#2}
\providecommand{\BIBentrySTDinterwordspacing}{\spaceskip=0pt\relax}
\providecommand{\BIBentryALTinterwordstretchfactor}{4}
\providecommand{\BIBentryALTinterwordspacing}{\spaceskip=\fontdimen2\font plus
\BIBentryALTinterwordstretchfactor\fontdimen3\font minus
  \fontdimen4\font\relax}
\providecommand{\BIBforeignlanguage}[2]{{%
\expandafter\ifx\csname l@#1\endcsname\relax
\typeout{** WARNING: IEEEtran.bst: No hyphenation pattern has been}%
\typeout{** loaded for the language `#1'. Using the pattern for}%
\typeout{** the default language instead.}%
\else
\language=\csname l@#1\endcsname
\fi
#2}}
\providecommand{\BIBdecl}{\relax}
\BIBdecl

\bibitem{mattson2004pathways}
M.~P. Mattson, ``Pathways towards and away from alzheimer's disease,''
  \emph{Nature}, vol. 430, no. 7000, pp. 631--639, 2004.

\bibitem{zeisel2020world}
J.~Zeisel, K.~Bennett, R.~Fleming \emph{et~al.}, ``World alzheimer report 2020:
  Design, dignity, dementia: Dementia-related design and the built
  environment,'' 2020.

\bibitem{de2020artificial}
S.~de~la Fuente~Garcia, C.~Ritchie, and S.~Luz, ``Artificial intelligence,
  speech, and language processing approaches to monitoring alzheimer’s
  disease: A systematic review,'' \emph{Journal of Alzheimer's Disease}, no.
  Preprint, pp. 1--27, 2020.

\bibitem{luz2017longitudinal}
S.~Luz, ``Longitudinal monitoring and detection of alzheimer's type dementia
  from spontaneous speech data,'' in \emph{2017 IEEE 30th International
  Symposium on Computer-Based Medical Systems (CBMS)}.\hskip 1em plus 0.5em
  minus 0.4em\relax IEEE, 2017, pp. 45--46.

\bibitem{campbell2021paralinguistic}
E.~L. Campbell, R.~Y. Mes{\'\i}a, L.~Doc{\'\i}o-Fern{\'a}ndez, and
  C.~Garc{\'\i}a-Mateo, ``Paralinguistic and linguistic fluency features for
  alzheimer's disease detection,'' \emph{Computer Speech \& Language}, vol.~68,
  p. 101198, 2021.

\bibitem{pastoriza2021speech}
P.~Pastoriza-Dominguez, I.~G. Torre, F.~Dieguez-Vide, I.~Gomez-Ruiz, S.~Gelado,
  J.~Bello-Lopez, A.~Avila-Rivera, J.~Matias-Guiu, V.~Pytel, and
  A.~Hernandez-Fernandez, ``Speech pause distribution as an early marker for
  alzheimer's disease,'' \emph{medRxiv}, pp. 2020--12, 2021.

\bibitem{bucks2000analysis}
R.~S. Bucks, S.~Singh, J.~M. Cuerden, and G.~K. Wilcock, ``Analysis of
  spontaneous, conversational speech in dementia of alzheimer type: Evaluation
  of an objective technique for analysing lexical performance,''
  \emph{Aphasiology}, vol.~14, no.~1, pp. 71--91, 2000.

\bibitem{luz2021detecting}
S.~Luz, F.~Haider, S.~de~la Fuente, D.~Fromm, and B.~MacWhinney, ``Detecting
  cognitive decline using speech only: The adresso challenge,'' \emph{medRxiv},
  2021.

\bibitem{yuan2020disfluencies}
J.~Yuan, Y.~Bian, X.~Cai, J.~Huang, Z.~Ye, and K.~Church, ``Disfluencies and
  fine-tuning pre-trained language models for detection of alzheimer’s
  disease,'' \emph{Proc. Interspeech 2020}, pp. 2162--2166, 2020.

\bibitem{syed2020automated}
M.~S.~S. Syed, Z.~S. Syed, M.~Lech, and E.~Pirogova, ``Automated screening for
  alzheimer’s dementia through spontaneous speech,'' \emph{INTERSPEECH (to
  appear)}, pp. 1--5, 2020.

\bibitem{balagopalan2020bert}
A.~Balagopalan, B.~Eyre, F.~Rudzicz, and J.~Novikova, ``To bert or not to bert:
  Comparing speech and language-based approaches for alzheimer's disease
  detection,'' \emph{arXiv preprint arXiv:2008.01551}, 2020.

\bibitem{guerrero2020word}
J.~S. Guerrero-Cristancho, J.~C. V{\'a}squez-Correa, and J.~R. Orozco-Arroyave,
  ``Word-embeddings and grammar features to detect language disorders in
  alzheimer’s disease patients,'' \emph{TecnoL{\'o}gicas}, vol.~23, no.~47,
  pp. 63--75, 2020.

\bibitem{mirheidari2018detecting}
B.~Mirheidari, D.~Blackburn, T.~Walker, A.~Venneri, M.~Reuber, and
  H.~Christensen, ``Detecting signs of dementia using word vector
  representations.'' in \emph{INTERSPEECH}, 2018, pp. 1893--1897.

\bibitem{orange1996conversational}
J.~B. Orange, R.~B. Lubinski, and D.~J. Higginbotham, ``Conversational repair
  by individuals with dementia of the alzheimer's type,'' \emph{Journal of
  Speech, Language, and Hearing Research}, vol.~39, no.~4, pp. 881--895, 1996.

\bibitem{pan2020improving}
Y.~Pan, B.~Mirheidari, M.~Reuber, A.~Venneri, D.~Blackburn, and H.~Christensen,
  ``Improving detection of alzheimer’s disease using automatic speech
  recognition to identify high-quality segments for more robust feature
  extraction,'' \emph{Proc. Interspeech 2020}, pp. 4961--4965, 2020.

\bibitem{kerz2020becoming}
E.~Kerz, Y.~Qiao, D.~Wiechmann, and M.~Str{\"o}bel, ``Becoming linguistically
  mature: Modeling english and german children’s writing development across
  school grades,'' in \emph{Proceedings of the Fifteenth Workshop on Innovative
  Use of NLP for Building Educational Applications}, 2020, pp. 65--74.

\bibitem{qiao2020language}
Y.~Qiao, D.~Wiechmann, and E.~Kerz, ``A language-based approach to fake news
  detection through interpretable features and brnn,'' in \emph{Proceedings of
  the 3rd International Workshop on Rumours and Deception in Social Media
  (RDSM)}, 2020, pp. 14--31.

\bibitem{strobel2020relationship}
M.~Str{\"o}bel, E.~Kerz, and D.~Wiechmann, ``The relationship between first and
  second language writing: Investigating the effects of first language
  complexity on second language complexity in advanced stages of learning,''
  \emph{Language Learning}, vol.~70, no.~3, pp. 732--767, 2020.

\bibitem{Christiansen2017}
M.~H. Christiansen and N.~Chater, ``Towards an integrated science of
  language,'' \emph{Nature Human Behaviour}, vol.~1, no.~8, Jul. 2017.

\bibitem{manning2014stanford}
C.~Manning, M.~Surdeanu, J.~Bauer, J.~Finkel, S.~Bethard, and D.~McClosky,
  ``The stanford corenlp natural language processing toolkit,'' in
  \emph{Proceedings of 52nd annual meeting of the association for computational
  linguistics: system demonstrations}, 2014, pp. 55--60.

\bibitem{klein2003accurate}
D.~Klein and C.~D. Manning, ``Accurate unlexicalized parsing,'' in
  \emph{Proceedings of the 41st Annual Meeting on Association for Computational
  Linguistics-Volume 1}.\hskip 1em plus 0.5em minus 0.4em\relax Association for
  Computational Linguistics, 2003, pp. 423--430.

\bibitem{devlin2018bert}
J.~Devlin, M.-W. Chang, K.~Lee, and K.~Toutanova, ``Bert: Pre-training of deep
  bidirectional transformers for language understanding,'' \emph{arXiv preprint
  arXiv:1810.04805}, 2018.

\bibitem{sun2019ernie20}
Y.~Sun, S.~Wang, Y.~Li, S.~Feng, H.~Tian, H.~Wu, and H.~Wang, ``Ernie 2.0: A
  continual pre-training framework for language understanding,'' \emph{arXiv
  preprint arXiv:1907.12412}, 2019.

\bibitem{collobert2011natural}
R.~Collobert, J.~Weston, L.~Bottou, M.~Karlen, K.~Kavukcuoglu, and P.~Kuksa,
  ``Natural language processing (almost) from scratch,'' \emph{Journal of
  machine learning research}, vol.~12, no. ARTICLE, pp. 2493--2537, 2011.

\bibitem{kalchbrenner2014convolutional}
N.~Kalchbrenner, E.~Grefenstette, and P.~Blunsom, ``A convolutional neural
  network for modelling sentences,'' \emph{arXiv preprint arXiv:1404.2188},
  2014.

\bibitem{kim2014convolutional}
Y.~Kim, ``Convolutional neural networks for sentence classification,'' 2014.

\bibitem{ma2015dependency}
M.~Ma, L.~Huang, B.~Xiang, and B.~Zhou, ``Dependency-based convolutional neural
  networks for sentence embedding,'' \emph{arXiv preprint arXiv:1507.01839},
  2015.

\bibitem{wolf-etal-2020-transformers}
\BIBentryALTinterwordspacing
T.~Wolf, L.~Debut, V.~Sanh, J.~Chaumond, C.~Delangue, A.~Moi, P.~Cistac,
  T.~Rault, R.~Louf, M.~Funtowicz, J.~Davison, S.~Shleifer, P.~von Platen,
  C.~Ma, Y.~Jernite, J.~Plu, C.~Xu, T.~L. Scao, S.~Gugger, M.~Drame, Q.~Lhoest,
  and A.~M. Rush, ``Transformers: State-of-the-art natural language
  processing,'' in \emph{Proceedings of the 2020 Conference on Empirical
  Methods in Natural Language Processing: System Demonstrations}.\hskip 1em
  plus 0.5em minus 0.4em\relax Online: Association for Computational
  Linguistics, Oct. 2020, pp. 38--45. [Online]. Available:
  \url{https://www.aclweb.org/anthology/2020.emnlp-demos.6}
\BIBentrySTDinterwordspacing

\bibitem{costello2018multilayer}
C.~Costello, R.~Lin, V.~Mruthyunjaya, B.~Bolla, and C.~Jankowski, ``Multi-layer
  ensembling techniques for multilingual intent classification,'' 2018.

\bibitem{oza2008classifier}
N.~C. Oza and K.~Tumer, ``Classifier ensembles: Select real-world
  applications,'' \emph{Information fusion}, vol.~9, no.~1, pp. 4--20, 2008.

\bibitem{alghanmi2020combining}
I.~Alghanmi, L.~Espinosa-Anke, and S.~Schockaert, ``Combining bert with static
  word embeddings for categorizing social media,'' 2020.

\bibitem{wolpert1992stacked}
D.~H. Wolpert, ``Stacked generalization,'' \emph{Neural networks}, vol.~5,
  no.~2, pp. 241--259, 1992.

\bibitem{rudin2019stop}
C.~Rudin, ``Stop explaining black box machine learning models for high stakes
  decisions and use interpretable models instead,'' \emph{Nature Machine
  Intelligence}, vol.~1, no.~5, pp. 206--215, 2019.

\bibitem{loyola2019black}
O.~Loyola-Gonzalez, ``Black-box vs. white-box: Understanding their advantages
  and weaknesses from a practical point of view,'' \emph{IEEE Access}, vol.~7,
  pp. 154\,096--154\,113, 2019.

\end{thebibliography}

\newpage
\FloatBarrier

\end{document}